# Computational Approaches to Measuring the Similarity of Short Contexts: A Review of Applications and Methods

**Ted Pedersen**
University of Minnesota, Duluth

**Abstract** Measuring the similarity of short written contexts is a fundamental problem in Natural Language Processing. This article provides a unifying framework by which short context problems can be categorized both by their intended application and proposed solution. The goal is to show that various problems and methodologies that appear quite different on the surface are in fact very closely related. The axes by which these categorizations are made include the format of the contexts (headed versus headless), the way in which the contexts are to be measured (first-order versus second-order similarity), and the information used to represent the features in the contexts (micro versus macro views). The unifying thread that binds together many short context applications and methods is the fact that similarity decisions must be made between contexts that share few (if any) words in common.

## 1. Introduction

A *short written context* consists of one to approximately 200 words of text that is presented to a human reader as a coherent source of information from which a conclusion can be drawn or an action taken. The exact upper limit on the number of words is not crucial, and the definition of word can be interpreted somewhat flexibly.[1] Short contexts are units of text that are no longer than a typical paragraph, and may correspond with a sentence, a passage, a verse, a phrase, an expression, a quote, an answer, a question, etc. depending on the application. Short written contexts may be extracted from longer documents such as articles, books, Web pages, etc. or they may be self-contained complete sources such as email messages, abstracts of technical articles, profiles on social network sites, job and product advertisements, news summaries, comments on products or articles, etc.

---

[1] In general *word* is used for ease of exposition, but it should be understood that *word* can include terms, phrases, collocations, abbreviations, etc. such as *President Bill Clinton, baseball bat,* or *AMA.*. Similarly, the methods described here could tokenize text into n-character sequences rather than space separated words and still function as presented.

Short written contents are crucial sources of information, especially in an online world that places a premium on display space and is inhabited by users who are intensively multi-tasking and have limited time and attention for any one information source. Identifying similar short contexts is the key to solving many problems in Natural Language Processing, and recognizing the underlying relationships between these problems and methodologies will allow for more rapid porting of solutions to new problems, domains, and languages as the need arises.

Take word sense discrimination and email categorization as examples of typical short context problems. While these are clearly related in that they are semantically oriented tasks, in fact these both can be approached with identical methods since the underlying goal is to recognize which contexts are similar to each other, in order to group the usages of the target word into senses and to group email messages into topic clusters.

The Distributional Hypothesis (Harris 1954) explains why so many different problems can be solved by recognizing which contexts are similar to each other. This hypothesis holds that words that are used in similar contexts will tend to have the same or related meanings. This same idea was memorably expressed by (Firth 1957) in the phrase: "You shall know a word by the company it keeps." More recently (Miller 1991) found that two words are similar to the degree that their contexts are similar; in effect showing that words that keep the same company are very similar or synonymous in meaning. From this previous work it follows that texts made up of similar words will tend to be about similar topics. Thus, large families of applications and methodologies can be based on recognizing when words occur in similar contexts, and when written contexts are made up of similar words.

Measuring short context similarity is a challenging problem, since short contexts rarely share many words in common. This is not true of longer texts, where similarity can often be assessed simply by determining how many words are shared or overlap between the contexts.

## 1.1 Similar Contexts

Similarity between concepts is well-defined, especially when concepts are organized in a hand-crafted ontology or dictionary such as WordNet.[2] Two concepts are said to be similar to the degree that they are connected via *is-a* relations. For example, a *cat* is similar to a *mouse* because both have a path through the hierarchy to "mammals", while a *cat* is more similar to a *lion* (than a *mouse*) since both have a path to "felines", which in turn has a path to "mammal", indicating it is a more specific concept.

---

[2] http://wordnet.princeton.edu/

However, similarity between short contexts that often include multiple concepts is less clearly defined. Obviously most short contexts will not have an entry in an ontology to which they can be mapped and measured relative to other such entries. Even taking a short context apart concept by concept and attempting to map each concept to an entry in an ontology to create a representation of the context is not likely to succeed on a wide scale because of the limited coverage of even the most extensive dictionaries and ontologies. In order to scale to a range of problems, domains, and languages, short context similarity measurement must not rely on hand-crafted knowledge sources such as ontologies or dictionaries. Fortunately, the Distributional Hypothesis provides a foundation for assessing *contextual similarity* based on evidence found in text.

The nature and degree to which two contexts are similar will depend on the problem being solved. For example, two email messages may be considered similar if they concern the same general topic such as "business" versus "personal life". However, the distinction between the different meanings of a target word that appears in multiple contexts is often more fine grained, such as the difference been *line* as a "queue" or as a "formation in sports". But, in the task of automatic language identification, two contexts are considered similar if they are expressed in the same natural language. As such there is considerable variation in the granularity of similarity judgments, ranging from detecting synonymy and near-synonyms to differentiating among languages.

Similarity can be interpreted on an absolute scale, where the similarity between two contexts is reported as a score and some conclusion is drawn based on that value. Otherwise, it can be viewed in more relative terms, where the goal is to simply determine which contexts are more like each other and cluster them together in a way that reflects this.

## 1.2 Types of Short Contexts

An important differentiating characteristic among short contexts is whether they are *headed* or *headless*. Short contexts that are headed have a target word specified that serves as the focus of the context. One way of visualizing a collection of headed contexts is to imagine a concordance where every line from a book or archive of news articles that contains a particular target word is extracted.[3] For example, suppose that every line from every article in the New

---

[3] Programs that create concordances are sometimes known as KWIC programs (Key Word in Context). The Unix command "grep" is also commonly used to create concordances or extract lines of text that contain a given target word.

York Times in 2004 that mentioned the target word *Stone Age* was extracted. Each of these lines would be treated as a short context, and would have a single occurrence of *Stone Age* designated as the target. If multiple occurrences of a target word occur in a line, typically a short context is created for each, so that each instance is the target in only one context.

Applications that rely on headed contexts often seek to organize *N* contexts so that the underlying similarities of the *N* instances of the target word are discovered. For example, given 1,000 short contexts that contain the target word *Stone Age*, the goal may be to discover the number of distinct senses in which it is used. It might be there are two, one that refers to the pre-historic era, and another that is a more colloquial expression referring to a belief or practice that is out of fashion.

Headless short contexts do not contain any specific target word, so they are suitable for problems where the goal is to make determinations about the overall contexts and not a specific word in the contexts. Thus, problems that rely on headed contexts tend to focus on micro levels of context around a particular target word (where utilizing the entire short context may not be critical to achieve the desired goal). However, headless problems have no basis for focusing on any particular part of the short context, and are not interested in making determinations about something internal to the context, but rather the overall short context.

### 1.2.1 Examples of Headed Contexts

To illustrate, there are three headed contexts shown below, each of which include the target word *line:*

*(1) Please stand in <u>line</u> over there.*
*(2) I wish the <u>line</u> would move a little faster.*
*(3) That darn fish took my <u>line</u> and my hook.*

Contexts of this form can be used for word sense discrimination, which determines the number of senses in which a target word is used in a given collection of headed contexts. As a human reader can discern, (1) and (2) are the sense of *line* as "queue", and (3) is *line* as "cable" or "cord".

Note that a target word can have different surface forms, for example:

*(4) <u>President Clinton</u> traveled to Mexico over the weekend.*
*(5) The guest list included <u>President Bill Clinton</u>.*
*(6) <u>Clinton</u> founded both Parliament and Funkadelic.*
*(7) <u>William Jefferson Clinton</u> was born in Hope, Arkansas.*

Contexts of this type can be used to carry out name discrimination, which determines how many people are referred to by the different forms of the target word. In this example there are 2 identities: (4), (5), and (7) refer to the 42$^{nd}$ President of the United States, and (6) refers to the musician George Clinton.

### 1.2.2 Examples of Headless Contexts

Headless contexts are clustered based on the overall gist of the context and not (directly) on the characteristics any individual component in the contexts. The following are examples of headless contexts:

> (8) *Sorry, I can't make the staff meeting due a personal conflict.*
> (9) *Please schedule some time with me on Friday to discuss your idea.*
> (10) *Happy Birthday John, many happy returns!*

Here the task is to differentiate how many different topics are discussed in these messages. A human reader can easily determine that there are two broad level distinctions: (8) and (9) refer to meetings, while (10) is a personal greeting.

Headless contexts are often somewhat longer than the examples above and could include abstracts of articles, summaries of news items, test questions, or product reviews at an online shopping site. They can also be created in many different ways from longer contexts. The contents of a textbook or article could be divided paragraph by paragraph, where each serves as a headless context. The first paragraph of each articles in a newspaper archive could be extracted and used as headless contexts. Verses from the Bible or ayats from the Qur'an could serve as headless contexts. In all of these cases the goal of finding the similarity of short headless contexts is to organize the contexts based on some aspect of their content.

## 2. Applications

There are a wide variety of problems that can be solved by identifying similar short contexts. The distinction between headless and headed contexts highlights the types of problems that can be represented. Headed contexts are focused on a specified target word within the contexts which is the object of whatever task is to be performed. Headless contexts do not have such a focus, and as such there can be a great deal of variation in the types of distinctions that are made between contexts. Headless contexts can be drawn from a very narrow domain, for example abstracts of articles form a molecular biology journal, or they can be very diffuse, as in the case of news summaries drawn from around the world. The degree of specificity in the contexts will affect the nature and granularity of the distinctions that can be made.

Applications that differentiate among short contexts can be divided into two categories. First, a short context can be measured to see which is most similar to a given reference sample (which is itself a short context) that serves as a gold-standard point of comparison. This is a pair-wise operation that scores the similarity between the context and the reference sample. Multiple comparisons may be made in order to identify the most similar context to the reference sample. Second, $N$ contexts can be grouped into $k$ clusters in order to identify which contexts are most similar to each other. While this often involves pair-wise comparisons, with each pair-wise operation a context is assigned to a cluster, and the process continues until all of the contexts are assigned to a cluster. There is no reference context in this case, and so rather than comparing their similarity to a given gold-standard, the goal is to group together the contexts that are most like each other, and least like the contexts in the other clusters.

## 2.1 Headless Applications

Headless short contexts occur naturally in many different forms and it isn't possible to enumerate all their variations. However, a few examples include email messages, encyclopedia entries, short answers on written tests, news bulletins, stock market updates, weather reports, product recommendations, blog entries, profiles on social network sites, and abstracts of technical articles. In addition, headless short contexts can be extracted from larger texts. For example, the first paragraph from a longer news article can provide a short context summary of the main facts and may be sufficient for deciding if the longer article should be retrieved.

New forms of short headless contexts are emerging due to advances in technology (among other reasons). For example, text messaging with SMS (short message services) is by its very definition a form of a short written headless context. Entire messages are expressed in less than 150 characters, which usually includes many abbreviations and shorthand expressions (LOL for laugh out loud, etc.) which can be viewed as words.

### 2.1.1 Pair-wise Comparison of Headless Contexts to Reference Samples

A reference sample is a short context that is used just like an answer key to grade a test. There are many tasks in Natural Language Processing where measuring the similarity to a reference sample is essential. For example, in authorship identification and plagiarism detection the goal is to determine if a piece of writing should be credited to the author represented by the reference sample.[4] While it is the case that writing samples used for these problems may sometimes be much longer than short contexts, they provide useful and intuitive examples of this type of application.

Automated NLP systems are often evaluated based on comparisons to a gold-standard reference sample created by a human expert, since in the end the goal of many NLP systems is to replicate human performance. Thus, evaluation of NLP systems is often a question of determining how similar the automatic output is to the human generated gold-standard. This is particularly relevant here, since in many cases the output of text generation, Machine Translation, and summarization systems are short contexts.

Many evaluation measures already exist for these tasks, and they tend to rely on string matching, that is determining if the system output has used the same words as the human reference sample. While a system that uses the same words (in approximately the same order) as a human reference sample is likely deserving of a high score, a system that makes different lexical choices and does not use many of the same words could still have output very similar to the gold standard, yet this would not be credited by evaluation measures that focus on string matching. This can be overcome to some degree by creating multiple human gold-standard answers, but in the end evaluation will still depend on a system producing the same words as the reference sample, and will not account for deeper similarities that may exist.

For example, BLEU (Papineni 2002) is a Machine Translation evaluation measure that finds phrasal matches (n-grams) of varying length. While it allows for the use of multiple reference samples, in the end a high score is only obtained by using the same words as appear in the reference samples. The METEOR evaluation measure (Banerjee 2005) also uses multiple reference samples, and it allows for some variation in lexical choice by considering the synonyms of words in the reference sample as matches in addition to direct matching.

Content filtering is another application where headless contexts are compared to reference samples. Web search results can be filtered to protect children from certain kinds of content by providing short context reference samples of inappropriate content, and filtering any Web search results that are overly similar. More ominously, these techniques could also be used to censor materials that are inconsistent with an officially stated government or company position. Web search engines return snippets, which are a kind of short context that is made up of a few lines of text that includes the search term, the Web page title, and the URL. It would be too time consuming to scan each and every Web page returned by a search engine, so short context comparisons are used in order to allow for speedy presentation of Web search results even when using such filters.

---

[4] In authorship identification the attribution of a piece of writing is unknown, and the goal is to determine if it should be credited to the author of the reference sample. In plagiarism detection, the attribution of a piece of writing is claimed, but may be re-assigned if it is overly similar that of the reference sample, which is attributed to a different author.

The profiles that people create and place on social networking sites such as Orkut, MySpace, or Facebook are often expressed as short headless contexts. Many people use these services to meet friends or find romance, and are hoping to meet people who share interests and tastes similar to their own. Thus, another possible application would be to automate matchmaking. A user who is searching for that special someone could provide their profile as a gold standard reference sample, and have that compared to candidate profiles to determine who might be a suitable match. Given the great variation in how people express themselves and describe their interests, methods that are based on direct matching will likely miss many possible candidates.

Automated grading or screening of short answers from students on written tests is another example of a pair-wise similarity comparison to a reference sample. Each student response can be compared to the correct answer, and those that pass a certain similarity threshold can be regarded as relevant and potentially good answers, while those that score poorly might be known to be badly off topic or irrelevant. A related task is generating appropriate feedback during an automated tutoring session, where a student response in a dialogue might be similar to one that is known to be an indicator of the need for a certain form of assistance that can then be provided by the automated tutor. Finally, automatic Question Answering systems seek to find the correct answer to a question posed by a user, and this can often involve identifying passages that are similar to the question which can then serve as the basis for generating an answer.

### 2.1.2 Clustering N Headless Contexts

Short context clustering is at the core of organizing content such as email or short news articles based on their overall topic or subject matter. Many of these contexts will not have many words in common with other contexts given their brevity and the wide range of lexical choice that exists even in specialized domains. In general clustering is an exploratory operation, and the goal is to discover how many groups of similar contexts are present in a collection. In practice the "correct" number of clusters is unknown, so that must be discovered as a part of the clustering process. One such method is the Adapted Gap Statistic (Pedersen 2006), which determines at which point adding to or reducing the number of clusters to which contexts are assigned does not result in any added improvement to the quality of the solution, and stops at that point.

Beyond the familiar examples of clustering email and news articles, there are other interesting applications that measure the similarity of headless contexts. For example, (Thabet 2005) employs what will be described as first-order similarity methods to cluster 24 of the 114 suras of the Qur'an. They found two clusters based on this analysis, each of which corresponds to the place of revelation of the sura (either Mecca or Medina). The 24 suras included in this study were greater in length than 1,000 words, which means they are not short headless contexts as defined here. However, the Qur'an consists of 6,236 ayats (verses) that make up the 114 suras, and each of those can be treated as a headless context. Thus, it would be possible to repeat this study and focus on the ayats as short headless contexts and see if the same types of clusters emerge as were discovered in the analysis of the suras.

It is important to note that clusters of headless contexts do not carry with them any type of label or tag that summarizes their contents, so strictly speaking the goal of clustering similar contexts is to determine which contexts belong together, and how many different groups are formed. Then a human can review the clusters and assign a suitable label or description. The ability to automatically label clusters with descriptive tags based on their content is an important problem, since requiring human intervention to label and interpret clusters is not practical on a large scale. One such method of automatic labeling is described in (Kulkarni 2005), where the most discriminating and descriptive two word sequences (bigrams) found in each of the clusters are used as tags.

The granularity of the resulting clusters of headless contexts will depend on the nature of the input. If the short contexts are abstracts of medical articles that relate to heart disease, then the resulting clusters will make distinctions within that domain. However, if the contexts are the contents of an email inbox, then the discovered clusters will likely be much broader.

## 2.2 Headed / Target Word Applications

Headed short contexts each contain a single occurrence of a specified target word. Common examples include concordances created by Key Word in Context (KWIC) programs, or Web search results, particularly when the user is searching for a single word or name. Applications that focus on target words seek to discriminate or disambiguate the senses of that word, and identify the instances of the target word that are used in each sense. This is achieved by grouping headed contexts into clusters, on the presumption that occurrences of the target word that occur in similar contexts will have similar or the same meanings.

### 2.2.1 Pair-wise Comparison of Headed Contexts to Reference Samples

(Lesk 1986) presents a method of word sense disambiguation that compares a short headed context to reference samples, where those reference samples are the dictionary definitions of the possible senses of the target word in the headed context. The similarity of the short context in which the target word occurs is measured against those reference samples, and the target word is assigned the sense with the definition that is most similar to the context in which it occurs. Similarity is measured by counting the number of words shared between the context and a definition (i.e., the Matching Coefficient). This is an example of a "bag of words" approach where the position of the words in the original context is not considered, and there is no phrasal matching. This method was extended in the Adapted Lesk Measure (Banerjee 2003) which measures the similarity between two concepts by determining how many words and phrases are shared between their definitions. In addition, it allows the definitions to be augmented to include the definitions of concepts that are closely related based on the structure of WordNet. This increases the size of the contexts and enhances the chances of finding matches between contexts.

There is an important distinction between word sense *disambiguation* and *discrimination*. Disambiguation assigns a word a meaning from a pre-existing sense inventory, usually provided by a dictionary or some other hand-crafted resource. This is the goal of Lesk's approach. Discrimination divides the contexts in which a target word occurs into clusters, each of which is made up of contexts that use that target word in a particular "sense". However, in discrimination there is no sense inventory available, and so the clusters do not have a definition or sense label associated with them, it simply determines how many different senses are used and which contexts belong to which sense cluster. Note that "senses" for a word as discovered by a discrimination method may be somewhat different than what is found in a dictionary since they will depend on the particular collection of contexts being clustered. Thus, disambiguation requires a dictionary or other hand-crafted resource that defines the possible senses. This is a significant constraint on methods that seek to be portable to new domains and languages, so as a practical matter most methods that rely on clustering and measuring the similarity of contexts will focus on discrimination.

### 2.2.2 Clustering N Headed Contexts

Automatic lexicography is an important application of measuring the similarity of short headed contexts. The overall goal is to discover the meanings of words based on the contexts in which they occur. This is typically divided into two steps: First, contexts that include the target word are grouped via a discrimination step, where each resulting cluster includes contexts that use the target word in very similar way (which is presumed to indicate they have a similar meaning). Second, after the contexts are clustered definitions are composed to explain the distinctions between them. Clustering short headed contexts carries out this first step but not the second. Automatically generating definitions of words based on the contexts that appear in each of the clusters remains a very important problem for future work.

Target words can be ambiguous like *line*, *interest*, or *bank*. They can also be names of entities or organizations, which are potentially just as ambiguous. For example, there are many people who have the name *George Miller*, including the father of WordNet, two different film directors, and a member of the U.S. Congress from California. Both names and words can be discriminated using identical methods (Pedersen 2005), since the key is determining which contexts are similar to each other, which does not depend on the nature of the target word.

Web search results are another type of headed short context, especially when the search is for a single word or name which acts like a target word. The goal of clustering Web search results is essentially the same as clustering words or names, that is to see what are the underlying meaning or identities of the target word. For Web search the objective is to make sure that the user finds pages that are relevant to their query and not confused by ambiguity with words or names.

Many other applications are possible since the target word does not need to be a single surface form. Morphological variations of the target word can be used as the focus of headed contexts (such as *line, lines, lined,* and *lining*). Examples (4) – (7) showed that different forms of names may be used as the target. Finally, even more flexibly the target word could in fact be a set of words. This might be useful in grouping different words together in order to ascribe more generic qualities or characteristics to something, as would be the goal of sentiment discovery or perception discovery. Suppose the following set serves as the target words : *pacifist, warrior, militant, hero, bumbler, president, imam.* Headless contexts including each word would be collected and placed into a single collection of *N* headed short contexts. These contexts would then be clustered to see how many clusters are formed, and which contexts are grouped together. This would make it possible to see if, for example, *president* appears in contexts similar to that of a *hero* or a *bumbler.*

Headed contexts can be very short and still provide sufficient evidence to make distinctions about the underlying sense or identity of the target word. (Choueka 1985) shows that observing two words to the left and two words to the right of a target word frequently provides sufficient context for humans to make disambiguation decisions.

## 3. Summary of Applications

Table 1 provides a summary of the different types of contexts and examples of applications and how they are formulated as problems.

**Table 1** Applications by type of contexts and problem formulation

|  | Headed Contexts (with target word) | Headless Contexts |
|---|---|---|
| Pair-wise measure of context to reference context / gold standard | *Disambiguation* of word senses (compare context to definition). | *Evaluation* of NLP systems for Machine Translation, generation, summarization. *Grading* short test answers. *Filtering* content. |
| Cluster *N* contexts into *k* groups | *Discrimination* of word senses, entity names, and Web search results. | *Categorization* of email, abstracts, news. |

Despite the differences among the applications mentioned above, all of them can be approached using methods that identify similar short contexts. In its simplest form, similarity can be identified by finding words shared between contexts. However, the absence of shared words is not sufficient grounds for concluding that two contexts are not similar. This poses a significant problem for short contexts, which because of their length may not show obvious surface similarities despite conveying nearly identical information. For example, the following contexts are nearly synonymous yet share no words in common:

> *(11) The Secretary of State gave a speech condemning the attack.*
> *(12) Yesterday's military strike was denounced by Condoleezza Rice.*

The fundamental question then is how can the similarity between contexts such as (11) and (12) be recognized and measured. In the following sections methods that identify similarity contexts are overviewed, starting with first order-methods, which are most suitable for contexts where at least some words are shared between them. Then second-order methods are introduced, which have some potential to solve problems like these.

## 4. First-order Similarity

First-order methods are the most intuitive approaches to measuring the similarity of short contexts. They rely on finding the number of words that are shared (or overlapping) between the contexts. Two contexts that share a large percentage of their words are likely to be similar, and this can be detected by simple string matching. The effectiveness of first order matching when contexts share many words can be seen in the following example:

> *(13) I visited Russia in 1996.*
> *(14) In 1996 I went to Russia.*

Contexts (13) and (14) are similar to the point of being nearly identical, and it's clear that the differences in word order have no impact on this judgment. The count of the number of words they share will depend on if a stop-list is used; this is simply a list of non-content words such as conjunctions, articles, etc. that occur in all kinds of contexts and are not considered useful for discriminating between contexts and are therefore removed prior to matching. In (13) and (14) *I, in,* and *to* would likely be removed as stop words, leaving *Russia* and *1996* as matches.

First-order similarity scores are based on the number of matching words. For example the Matching Coefficient reports the number of shared words, while the Jaccard and Dice Coefficients scale the number of matches by the length of the contexts. Regardless of the measure used, the key piece of information for first-similarity scores is the number of shared or overlapping words. These methods all treat the contexts as "bags of words" and do not consider the position of words when finding matches. This type of matching has been extended in various measures of first-order similarity by giving extra credit for matches that are longer than one word (e.g., (Papineni 2002), (Banerjee 2003)). This is based on the premise that matches of more than one word are significantly rarer and more informative than just one word matches.

First-order similarity can also be measured using the vector space model. Each word in a collection of contexts is treated as a feature, and each context is represented by a vector that shows which of these features occur within it. Feature values can be binary and indicate if the feature occurs or not, or the frequency of the feature in the context can be shown. Then, two contexts that share words in common will both have non-zero values for their associated features in the vector space model. Once these vectors are constructed the similarity between pairs of contexts can be measured using the cosine measure, and clusters of contexts can be formed based on these scores if desired. While there are many possible clustering algorithms, in general they all rely on similarity scores like the cosine to determine which contexts are most like each other and belong in the same cluster, and which should be separated into different clusters.

First-order similarity can be effective for longer contexts and full length documents, especially when the objective is to make broad topic level distinctions. In general for longer texts there will be some number of shared words, however, lack of consistent vocabulary and variations in lexical choice can still be a problem. First-order methods can also perform well with short contexts; however this requires that the vocabulary be somewhat regular or standardized, as would be the case with abstracts of technical articles, or weather and stock market reports. For many other short contexts even slight variations in terminology and lexical choice can leave little or no information for such methods.

A further complication with short contexts is that similar contexts may not be of similar lengths. A Question Answering or Intelligent Tutoring system must determine if a response to a question is relevant, and in general questions and answers can be of very different lengths. This is similar to the problem of passage retrieval, where the goal is to identify a passage of text that is similar to or satisfies a user query. Both the query and the passages to be retrieved are kinds of short contexts, they are just of different length and structure. The query terms represent an extremely short context, especially since a user is not likely to specify all the possible variations of their query. For example a user may mention *oil* but not *petroleum* in a query. This may be addressed via Query Expansion (Salton 1971), which automatically adds related and synonymous words to a query, so that it has a better chance of finding matches.

In the case of (12) and (13), *visited* could be expanded to *went* (and vice versa) on the basis of synonym or near-synonym entries in a dictionary or other lexical resource. If this expansion was made, then these two contexts would match exactly despite differences in word order and lexical choice. While hand-crafted resources such as dictionaries and thesauruses are invaluable for these operations, they may lack coverage and not scale to new domains or languages. As a result there is a long history of work in automatically creating thesauruses by clustering words from large corpora (e.g., (Crouch 1992), (Qiu 1993)).

Beyond expanding words in the context, it may be possible to increase the effectiveness of first-order similarity matching by augmenting a context with related text (assuming that such texts can be identified, since this is in fact an instance of our original problem of measuring the similarity of contexts). One method of augmentation is found in the Adapted Lesk Measure (Banerjee 2003), which measures first-order similarity between definitions of concepts found in WordNet to measure their relatedness. Dictionary definitions tend to be very short and are often hard to match since they contain few shared words. However, the Adapted Lesk Measure overcomes this by utilizing the networked structure of WordNet so that a definition is augmented to include other definitions that are directly linked via a WordNet relation (*is-a, is-a-part-of,* etc.). This increases the size of the contexts considerably, and improves the chances of finding direct first-order matches.

Finally, another possible enhancement to first-order similarity is to loosen the matching criteria, and allow for fuzzy or partial matches. This can be achieved by matching n-character sequences rather than entire words, or using edit distances, where words that only differ by a few edit operations are considered similar. For example, *industry* and *industrial* would not match directly, but *industrial* can be transformed via three edit operations into *industry* (remove the *a* and the *l* and change the final *i* to *y*). Edit distance can be problematic however, in that there are many words that differ by just a few edit operations and yet have completely different meanings (e.g., *cat* and *hat*, *abduction* and *deduction*, etc.) It might be more reliable to stem the words in the contexts to reduce them to their morphological base forms (e.g., reducing *digging, digs*, and *dug* to the stem *dig)*. However, stemmers are language dependent and may suffer from limitations in coverage especially for certain domains, and they may simply not be available for many languages.

First-order similarity is a reasonable option for many kinds of texts, although it has limitations for short contexts since there are often few shared words. This can be addressed by stemming the contexts, expanding words with synonyms or other related words, augmenting the contexts with additional related texts, or by employing fuzzy matching techniques. However, stemming is a language and domain dependent operation, query expansion requires some knowledge of the intended sense of a word (e.g., if *bank* is the query term, should it be expanded to *shore* or *credit union*?), context augmentation requires the ability to identify similar contexts in the first place, and fuzzy matching can sometimes be too generous and introduces false similarities.

First-order similarity measures might best be viewed as appropriate for applications where identifying direct matches is a priority, as would be the case in plagiarism detection. However, in many other cases direct matching is simply too restrictive and is unable to detect many more subtle instances of similarity. First-order methods will not suffer from many false positives; if two contexts share many words then they are likely similar. However, there will be many false negatives. Many short contexts that are similar do not share many words, but first-order methods will not detect this.

## 5. Second-order Similarity

S*econd-order similarity* measurement tries to achieve some of the same goals as word expansion, context augmentation, and fuzzy matching by replacing the contexts with some other derived representation that is richer in content and provides a more substantial basis for measuring similarity.

Second-order similarity can be introduced by considering the problem of measuring the similarity of single word contexts. While this might seem like an unusual application, in fact it is quite common to group similar words together. Word clustering is the basis for Query Expansion, among many other possible tasks. Suppose that the contexts are as follows:

*(15) score*
*(16) goal*

Obviously direct first-order matching of the two contexts will fail, and edit distance or other fuzzy matching techniques will not provide anything useful. When lacking sufficient information for first-order methods, the next step is to consider a second-order similarity method. In general the idea is to replace the context with something else that will still represent it, and yet hopefully provide more information from which similarity judgments can be made.[5] Assuming the availability of a dictionary like WordNet, it would be possible to replace a word with its definition; *score* is replaced with (17) and *goal* with (18):

*(17) a number that expresses the accomplishment of a team or an individual in a game or contest*
*(18) a successful attempt at scoring*

While the contexts are now much larger, they are still short contexts and in this case there are still   no shared words between the definitions. However, note that (18) uses a form of *score*, so if the contexts were expanded to include both the definition and the original word then some similarity *might* be detected via edit distance or fuzzy matching of the contexts. However, it should be clear that expanding with dictionary definitions does not completely solve the problem and in fact just brings us back to a variation of the original Lesk problem.[6]

Recall that the Adapted Lesk Measure addresses the problem of short contexts by augmenting them with the definitions of related words. This could be done for (17) and (18), and could even be taken one step further, and each word in the definitions could be replaced with their definitions or some other representation. In fact, this is exactly the approach of (Patwardhan 2006) in creating the vector measure for WordNet-Similarity, which is an example of a second-order similarity method. It augments the definitions as in Adapted Lesk, and then replaces the words in the definitions with word vectors that provide co-occurrence information about each word.[7]

---

[5] We introduce a distinction between expand and replace. In expansion we assume that the original word remains as a candidate for matching, while in replacement the original word is not a part of the matching process.

[6] The point of this example is not to advocate for the replacement of words by dictionary definitions, but rather to give an intuitive idea of what it means to replace a word in a second-order method. In fact the use of a hand-crafted dictionary is problematic in that it will be limited in coverage and methods that rely on such resources will not easily scale to new domains and languages.

## 5.1 Word Vectors

Hand-crafted resources like WordNet will not have sufficient scope and coverage for generic solutions to the problem of matching similar contexts. Thus, second-order methods must rely on more readily available sources of information with which to expand or replace words in short contexts if the methods are to scale to new domains and languages. The most generic and flexible option would be to derive that information from large corpora or the contexts themselves, which is the basis of the approach to be described here. In particular, words will be replaced by vectors that represent the contexts in which that word occurs, again following the Distributional Hypothesis that words that occur in similar contexts will have similar meanings, and that contexts made up of similar words will be similar to each other.

The context in which a word occurs has both a micro and a macro view. The micro view is made up of the words that surround or co-occur with a particular word (keeping it company), and the macro view corresponds to the collection of contexts in which a word occurs. The micro view is sometimes referred to as local context, while the macro view may be referred to as topical context.

### 5.1.1 Micro View of Context

The micro view of the context in which a word occurs is represented by a vector that indicates the words that co-occur in relatively close proximity, often within 5 or 10 positions of each other, although the actual size of this window can vary considerably. The values in the micro context vector can take many forms. They may be frequency counts that show how often the two words occur together, or binary values that indicate whether they do or not. These can also be measures of association between the word and each of its co-occurrences, indicating how strongly associated they are, in order to separate those words that occur together just by chance from those that have some systematic relationship. The micro view might be based on the co-occurrence behavior of the word in one very large corpus where it occurs many times, or collected across hundreds of short contexts where it only occurs once. The micro-vector only shows the co-occurring words, it does not indicate where those co-occurrences may have been observed.

---

[7] WordNet-Similarity is a freely available software package that includes a number of measures of semantic similarity based on the lexical database WordNet, including the Adapted Lesk measure and the vector measure. It can be downloaded from http://wn-similarity.sourceforge.net.

The co-occurrence counts or association values that make up a word vector based on micro context can be obtained locally from the contexts that are being measured for similarity, or globally from some external corpora that is independent of the shorts contexts to be represented. For example, suppose there are 10 headed short contexts that contain the target word *cricket* and are to be clustered based on their similarity (perhaps to determine if they pertain to the insect or the sport of the same name). In all likelihood the amount of co-occurrence data that can be obtained from just 10 short contexts is very small, so a global approach that draws co-occurrence information from larger external corpora would be advisable. In this case one possibility would be that co-occurrence information for all the words that appear in the short contexts of *cricket* could be obtained globally from a large sample of newspaper text. If however there are 10,000 short contexts, it might be possible to obtain sufficient co-occurrence data (in the form of counts or association scores) from the data itself, thus suggesting a local approach. Some tasks will inherently have only a small number of contexts available, such as those that compare a context to a reference sample. In those cases a global method for obtaining co-occurrence data is almost always required.

In addition to drawing upon corpora for co-occurrence data for micro context vectors, the Web also presents an abundant and readily available global source of co-occurrence data. Each word from a collection of short contexts could be used as a search query to the Web either in whole or to some specific part of it such as Wikipedia, Google News, etc. The pages returned from that query would then be used to derive the co-occurrence data for that word (e.g., (Sahami 2006))

Regardless of where the co-occurrence information comes from, a word vector based on micro context will show the company a word keeps, and uses this as a unique representation of that word. Words that have very similar co-occurrences are judged to be similar since they keep the same company, and perhaps could be substituted for each other (a common test of synonymy). Note that there is some granulation in how similar such words are, and that depends on the size of the window used to determine co-occurrence. If any word that occurs within 50 positions of a word is considered a co-occurrence, then quite a few words may show signs of similarity, perhaps at a topic level. However, if a co-occurrence must occur within two positions of the word, then the similarity is more likely to approach synonymy.

The micro view of context has been widely used in the Word Sense Disambiguation literature (c.f., (Ide 1998)). Both (Pedersen 1997) and (Purandare 2004) have shown the micro contexts provide sufficient information for good results in automatic word sense discrimination.

### 5.1.2 Macro View of Context

The macro view of context adjusts the maxim about judging a word based on the company it keeps. Instead, it holds that : "you can judge a word by the places where it is seen." For example, the word *polymer* will tend to be restricted to texts about molecular structure or science, whereas the word *be* will appear in nearly any written text. Most words are of course somewhere in the middle of this spectrum.

Macro contexts can be obtained locally from the contexts that are being clustered, or from some other global external source. If there are 10,000 short contexts to be clustered, then it may be that simply recording in which of these contexts each word occurs will provide sufficient information to make a discrimination decision. However, if the task is to discriminate among 10 instances of *cricket* or to compare a single context to a few reference samples, then a global collection of contexts external to the problem should be used. Possible sources of macro contexts could include textbooks or encyclopedias, where each paragraph is treated as a unit of context in which a word may be seen.

The macro view of context has been used extensively in Latent Semantic Analysis (Landauer 1997), where it has been applied to problems in automated essay grading, synonym identification, and short passage similarity measurement, among many others. (Levin 2006) uses the macro view of context for word sense discrimination, and reports encouraging results.

The distinction between the micro and macro view of context is subtle, in that words that occur in specific types of macro contexts will likely occur in specific micro contexts as well. However, a word with multiple senses might have very different friends depending on the places it is seen, that is to say the word might have different circles of friends when it is used in different senses. When *interest* is used in contexts about business it might well refer to money paid for the use of money, and be surrounded by words like *rate* and *mortgage,* whereas when used in more general contexts it could be surrounded by anything a person is interested in, such as *art, music, sports,* etc. Micro context captures the localized behavior of a word, while the macro context in which a word occurs may make a more general topic level distinction.

A word can be represented by a vector that shows the words it co-occurs with (micro), or a vector that shows the contexts in which it occurs (macro). In either case, these word vectors are used to create a representation of the short contexts that is more amenable to measuring similarity between short contexts than first-order methods.

## 5.2 Context as an Average of Word Vectors

Short contexts normally consist of more than one word, so even after a word vector has been created based on the micro or macro view of context, there is still the question how to represent an entire short context, not just individual words. Since each of the words in a context is replaced with a vector, a natural solution is to represent the context with the centroid of the word vectors. Then, contexts can be measured for similarity using the cosine or any other vector based measure of similarity. The centroid is a vector that has the same dimensions as the word vectors, and is equally distant from all of them, and is therefore located in the middle of the word vectors. Mathematically this is accomplished by finding the average of the vectors, so the centroid is simply an averaged word vector for that context.

Averaging word vectors to represent a context has been used with the macro view of context in Latent Semantic Indexing (Deerwester 1990) and Latent Semantic Analysis (Landauer 1997). Averaging word vectors based on the micro view of context has been pursued by (Schütze 1998) and (Purandare 2004) for carrying out word sense discrimination, and (Patwardhan 2006) for measuring semantic relatedness between concepts.

Prior to creating the averaged vectors to represent the contexts, the micro view of context results in the construction of a word by word matrix, where the rows of this matrix represent the words in the contexts, and the columns represent the co-occurring words that may be obtained locally (from the short contexts themselves) or globally (from other resources like corpora or the Web). The values in the cells of this matrix may contain binary values that indicate if the words have co-occurred, frequency counts showing how often they have occurred together, or measures of association that show how dependent they are on each other. The macro view results in a word by context matrix, where the rows represent the words that occur in the context. The columns represent the contexts in which the words have occurred, and these may be obtained locally (from the contexts themselves) or globally (from some other resource outside of the contexts). The values in the matrix cells can be binary to indicate that a word occurred in a context, or they can include a frequency count that shows how many times a word occurred in a context. In both cases the resulting matrix will be very sparse, and it may be beneficial to perform dimensionality reduction on the either the word by word matrix (micro view) or the word by context matrix (macro view) to reduce the sparsity and dimensionality of the matrix, and smooth the values in the cells so that there are fewer zero valued cells. This will make it possible to draw finer grained distinctions between contexts when measuring their cosine. Dimensionality reduction is typically performed via Singular Value Decomposition (SVD), which has the effect (simply put) of trying to group together the columns in these matrices that are similar to each other, and reduce the dimensionality of the matrix by compressing similar columns together. SVD is an optional step however, and is not required to create the context representations from the averaged word vectors.

The resulting word matrices are used to create the context representations, and this proceeds very simply. Each word in a context is replaced with a word vector (that has been optionally reduced via SVD). If there is no word vector available, then the word is dropped from the context. Once all the word are replaced, then the word vectors are averaged to find the centroid, and that centroid represents the context. This is done for every context of interest, so there is one vector per context. These can be clustered or measured for pair-wise similarity. Clustering and measuring the similarity of vectors are well studied areas, and many alternative methods exist. For short context problems the important point is not how the clustering or pair-wise measurement will be done, but rather how the short context representations can be created to allow for similarity judgments to be made even when there are few shared words between the contexts.

## 6. Conclusion

This article describes how short contexts can be represented to allow similarity measurements to be taken between them, even when short contexts may share few (if any) words in common. For short contexts, first-order methods may not find enough words in common between the contexts to proceed. If this is the case, then second-order methods provide a viable alternative, where each word in a context is replaced by a vector that represents the contexts in which that word occurs. The view of context can either be based on locally surrounding co-occurrences (micro) or the contexts in the word occurs (macro). The word vectors are averaged together to create a centroid that represents the context. The use of word vectors to replace words in contexts allows for information to represent the words be obtained from other sources such as very large corpora or the Web, and thereby overcome the relative scarcity of information present in the short contexts.

## Acknowledgements

The open-source software package SenseClusters supports clustering headless and headed contexts using both first-order and second-order similarity. Second order similarity is measured using micro or macro views of context. It can also create and cluster word vectors. It is freely available from the following URL: http://senseclusters.sourceforge.net. SenseClusters was implemented at the University of Minnesota, Duluth by Amruta Purandare (2002-2004) and Anagha Kulkarni (2004-2006), with support for Latent Semantic Analysis being added by Mahesh Joshi in the summer of 2006.

# References

Banerjee, S. and Lavie, A. 2005: 'METEOR : An Automatic Method for MT Evaluation with Improved Correlation with Human Judgments'. In *ACL 2005 Workshop on Intrinsic and Extrinsic Evaluation Measures for MT and/or Summarization*. Ann Arbor, MI: pp. 65-73.

Banerjee, S. and Pedersen, T. 2003: 'Extended Gloss Overlaps as a Measure of Semantic Relatedness'. In *18th International Joint Conference on Artificial Intelligence*. Acapulco, Mexico: pp. 805-10.

Choueka, Y. and Lusignan, S. 1985: 'Disambiguation by Short Contexts'. *Computers and the Humanities*, 19, pp. 147-57.

Crouch, C.J. and Yang, B. 1992: 'Experiments in Automatic Statistical Thesaurus Construction'. In *15th ACM-SIGIR Conference on Research and Development in Information Retrieval*. Copenhagen, Denmark: pp. 77-88.

Deerwester, S. and Dumais, S. T. and Furnas, G. and Landauer, T. and Harshman, R. 1990: 'Indexing by Latent Semantic Analysis'. *Journal of the American Society for Information Science*, 41, pp. 391-407.

Firth, J. R. 1957: 'A Synopsis of Linguistic Theory, 1930-1957'. In *Special volume of the Philological Society*. Oxford: Blackwell.

Harris, Z. 1954: 'Distributional Structure'. *Word*, 10, pp. 146-62.

Ide, N. and Véronis, J. 1998: 'Word Sense Disambiguation : The State of the Art'. *Computational Linguistics*, 24, pp. 1-40.

Kulkarni, A. and Pedersen, T. 2005: 'Name Discrimination and Email Clustering using Unsupervised Clustering and Labeling of Similar Contexts '. In *Second Indian International Conference on Artificial Intelligence*. Pune, India: pp. 703-22.

Landauer, T. and Dumais, S. 1997: 'A Solution to Plato's Problem: The Latent Semantic Analysis Theory of Acquisition, Induction and Representation of Knowledge'. *Psychological Review*, 104, pp. 211-40.

Lesk, M. 1986: 'Automatic Sense Disambiguation Using Machine Readable Dictionaries: How to tell a pine cone from a ice cream cone'. In *SIGDOC-86: 5th International Conference on Systems Documentation*. Toronto, Canada pp. 24-26.

Levin, E. and Sharifi, M and Ball, J. 2006: 'Evaluation of Utility of LSA for Word Sense Discrimination'. In *Proceedings of the Human Language*

*Technology Conference and the Sixth Annual Meeting of the North American Chapter of the Association for Computational Linguistics*. New York City: pp. 77-80.

Miller, G. and Charles, W. 1991: 'Contextual Correlates of Semantic Similarity'. *Language and Cognitive Processes*, 6, pp. 1-28.

Papineni, K., Roukos, S., Ward, T., and Zhu, W. J. 2002: 'BLEU: A Method for Automatic Evaluation of Machine Translation'. In *40th Annual meeting of the Association for Computational Linguistics*. Philadelphia, PA: pp. 311-18.

Patwardhan, S. and Pedersen, T. 2006: ' Using WordNet Based Context Vectors to Estimate the Semantic Relatedness of Concepts'. In *EACL 2006 Workshop Making Sense of Sense - Bringing Computational Linguistics and Psycholinguistics Together*. Trento, Italy: pp. 1-8.

Pedersen, T. and Bruce, R. 1997: 'Distinguishing Word Senses in Untagged Text '. In *Proceedings of the Second Conference on Empirical Methods in Natural Language Processing* Providence, RI: pp. 197-207.

Pedersen, T. and Kulkarni, A. 2006: ' Automatic Cluster Stopping with Criterion Functions and the Gap Statistic '. In *Demonstration Session of the Human Language Technology Conference and the Sixth Annual Meeting of the North American Chapter of the Association for Computational Linguistics*. New York City: pp. 276-79.

Pedersen, T., Purandare, A. and Kulkarni, A. 2005: 'Name Discrimination by Clustering Similar Contexts'. In *Sixth International Conference on Intelligent Text Processing and Computational Linguistics*. Mexico City: pp. 220-31.

Purandare, A. and Pedersen, T. 2004: 'Word Sense Discrimination by Clustering Contexts in Vector and Similarity Spaces'. In *Conference on Computational Natural Language Learning*. Boston, MA: pp. 41-48.

Qiu, Y. and Frei, H.P. 1993: 'Concept Based Query Expansion'. In *16th ACM-SIGIR Conference on Research and Development in Information Retrieval*. Pittsburgh, PA: pp. 160-69.

Sahami, M. and Heilman, T. 2006: 'A Web-based Kernel Function for Measuring the Similarity of Short Text Snippets'. In *Fifteenth International World Wide Web Conference*. Edinburgh, Scotland: pp. 377-86.

Salton, G. 1971: *The SMART Retrieval System : Experiments in Automatic Document Processing*. Englewood Cliffs, NJ: Prentice Hall.

Schütze, Hinrich 1998: 'Automatic Word Sense Discrimination'. *Computational Linguistics*, 24, pp. 97-123.

Thabet, N. 2005: 'Understanding the Thematic Structure of the Qur'an: An Exploratory Multivariate Approach'. In *Student Research Workshop at the 43rd Annual Meeting of the Association for Computational Linguistics*. Ann Arbor, MI: pp. 7-12.